\pdfoutput=1

\documentclass[11pt]{article}

\usepackage[final]{acl}

\usepackage{times}
\usepackage{latexsym}
\usepackage{multirow}
\usepackage{amsmath,amsfonts,amssymb}
\DeclareMathAlphabet{\mathbbold}{U}{bbold}{m}{n}
\usepackage{colortbl}
\usepackage[dvipsnames]{xcolor}
\usepackage[caption=false]{subfig} 
\usepackage{tcolorbox}
\usepackage{hyperref}
\usepackage{float}
\usepackage{tikz}
\usepackage[table,xcdraw]{xcolor}
\usepackage{enumitem}
\usepackage[T1]{fontenc}

\usepackage[utf8]{inputenc}

\usepackage{microtype}

\usepackage{inconsolata}

\usepackage{graphicx}
\usepackage{booktabs}
\usepackage[capitalize,nameinlink,noabbrev]{cleveref}
\usepackage[normalem]{ulem} 
\definecolor{myblue}{HTML}{1b62a5}
\definecolor{mygreen}{HTML}{279321}
\definecolor{mypurple}{HTML}{814faf}
\definecolor{myred}{HTML}{ca111d}
\definecolor{myorange}{HTML}{fd6b10}

\title{From Confidence to Collapse in LLM Factual Robustness}

\author{Alina Fastowski \hspace{.5cm} Bardh Prenkaj \hspace{.5cm} Gjergji Kasneci \\ \\
Technical University of Munich \\
\texttt{\{name.surname\}@tum.de}}

\begin{document}
\maketitle
\begin{abstract}
Ensuring the robustness of factual knowledge in LLMs is critical for reliable applications in tasks such as question answering and reasoning. However, existing evaluation methods predominantly focus on performance-based metrics, often investigating from the perspective of prompt perturbations, which captures only the externally triggered side of knowledge robustness. To bridge this gap, we introduce a principled approach to measure \textbf{factual robustness} from the perspective of the generation process by analyzing token distribution entropy in combination with temperature scaling sensitivity. 
These two factors build the Factual Robustness Score (FRS), a novel metric which quantifies the stability of a fact against perturbations in decoding conditions, given its initial uncertainty. To validate our approach, we conduct extensive experiments on 5 LLMs across 3 closed-book QA datasets (SQuAD, TriviaQA, and HotpotQA). We show that factual robustness varies significantly -- smaller models report an FRS of $0.76$, larger ones $0.93$ -- with accuracy degrading by $\sim$$60\%$ under increased uncertainty. These insights demonstrate how entropy and temperature scaling impact factual accuracy, and lay a foundation for developing more robust knowledge retention and retrieval in future models. We release our code at \url{https://github.com/afastowski/frs}.

\end{abstract}

\section{Introduction}  

Large Language Models (LLMs) have revolutionized natural language understanding and generation, demonstrating remarkable capabilities across tasks such as question answering, knowledge-intensive text generation, and reasoning \citep{Petroni2019Knowledge, Roberts2020HowMuch,hu2023large}. However, despite their strong performance, factual stability remains an open challenge. While a model may provide a correct answer under one set of decoding conditions, it may fail to do so when faced with minor perturbations, such as variations in sampling temperature. This raises a fundamental question: \textbf{how robustly is factual knowledge embedded in LLMs?  }

Traditional evaluations of factual knowledge in LLMs rely on accuracy-based metrics, which measure correctness under fixed conditions \citep{2017arXivtriviaqa, kwiatkowski2019natural, hendrycks2020measuring}. However, these methods do not account for uncertainty in fact retrieval -- whether a model is inherently confident in its answer or if its correctness is fragile under perturbations. Additionally, prior techniques often focus on single-temperature evaluations, failing to systematically analyze how factual outputs degrade as uncertainty is introduced \citep{Maynez2020Faithfulness, gabriel2021go}. To bridge this gap, we introduce a novel perspective: factual robustness, which examines not just whether an answer is correct but how resistant it is to internal uncertainty and temperature-induced variability.  

To quantify this, we propose the Factual Robustness Score (FRS), a new metric designed to measure the stability of factual knowledge within an LLM. Unlike previous work that assesses model performance at isolated temperature values, FRS integrates both entropy and temperature, providing a comprehensive robustness assessment. Entropy quantifies the model's intrinsic confidence in an answer, while the so-called ``breaking'' temperature is the first temperature greater than zero, at which the generation produces an incorrect answer. In other words,  it captures how much uncertainty the generation process can withstand, before shifting from the correctly produced answer to an incorrect one. By combining these two dimensions, FRS moves beyond simple accuracy and provides a deeper insight into how reliably knowledge is stored and retrieved within LLMs.  

Through extensive experiments across multiple LLM architectures and datasets, we show that higher-entropy facts degrade more under temperature perturbations, while lower-entropy facts remain more stable. Model size alone does not dictate robustness, as architectural and training differences also play a role. Additionally, factual stability varies by knowledge type, with numerical facts proving more resilient than others.


\noindent Our contributions are multifold:  
\begin{enumerate}[noitemsep,topsep=0pt]
    \item \textbf{We introduce the Factual Robustness Score (FRS)}, the first metric to systematically assess the stability of factual knowledge in LLMs by integrating entropy and breaking temperature.  
    \item \textbf{We analyze the impact of temperature on factual accuracy}, showing that increasing temperature systematically degrades correctness, but that this effect varies across models and knowledge types. 
    \item \textbf{We provide empirical evidence that factual robustness is not solely determined by model size}, highlighting the role of architecture and training methods in knowledge stability. 
    \item \textbf{We identify variations in robustness across different types of factual knowledge}, showing that certain categories, such as numerical facts, are more resilient than others.
    \item  \textbf{We demonstrate that factual robustness cannot be inferred from accuracy alone}, reinforcing the need for stability-focused evaluation methods beyond traditional correctness assessments. 
\end{enumerate}

\section{Preliminaries}\label{sec:preliminaries}

\textbf{Entropy} is a measure of disorder and randomness in a system. In text generation, it measures the spread of the probability distribution over possible tokens, with higher entropy indicating more randomness, and lower entropy reflecting more confident, deterministic predictions (see \cref{eq:entropy}). Following the intuition of entropy, we consider it as the first factor in our robustness assessments.
\begin{equation}\label{eq:entropy}
    H = -\sum_i{P(x_i) \log P(x_i)}.
\end{equation}%
\noindent\textbf{Temperature} is a parameter that controls the sharpness of the probability distribution during text generation. It operates by scaling the logits before applying the softmax, as defined in \cref{eq:temperature_computation}:
\begin{equation} P(x_i;t) = \frac{\exp\left(\frac{\log P(x_i)}{\max(t,\varepsilon\cdot\mathbbold{1}_{t=0})}\right)}{\sum\limits_{j} \exp\left(\frac{\log P(x_j)}{\max(t,\varepsilon\cdot\mathbbold{1}_{t=0})}\right)} \label{eq:temperature_computation}, \end{equation}%
where $t \geq 0$ is the temperature value, $\varepsilon > 0$ is a small number approaching $0$, in our experiments set to $10^{-4}$, to avoid zero divisions, and $\mathbbold{1}_{t=0}$ is the indicator function with the condition $t=0$. A lower $t$ makes the probability distribution more peaked, reinforcing the dominance of high-probability tokens and making the model more deterministic. Conversely, a higher temperature flattens the distribution, increasing the likelihood of selecting lower-probability tokens and promoting more diverse generations. The effect of temperature serves as the second factor in our robustness analysis.


\noindent\textbf{LLMs' True Factual Knowledge.}
To assess factual robustness in LLMs, we first identify facts the model answers correctly at \( t=0 \), where the probability distribution is sharply peaked, ensuring the most likely tokens dominate (see \cref{fig:modified_temp_certainties}). If the model has high confidence in the correct answer, it will always produce it at \( t=0 \).  However, as we increase the temperature towards \( t=1 \), the probability distribution flattens, allowing lower-probability tokens to be chosen. At \( t=1 \), temperature scaling no longer affects the probability computation (see \cref{eq:temperature_computation}), meaning we observe the model's intrinsic token distribution—its true knowledge representation, without artificially enforced certainty.  

Our key idea is as follows: if an initially correct answer starts shifting to an incorrect one as we increase the temperature, the fact is not robustly stored in the model. In other words, if the model can only produce the correct answer when artificially forced into a low-uncertainty setting (\( t=0 \)), then the fact is not stably embedded in its knowledge base. A truly robust fact should remain correct across a range of temperatures, reflecting consistent and confident knowledge retention. Furthermore, the initial entropy of a generated answer gives another factor about its robustness: if an answer is produced at high uncertainty, it is less robust from the beginning. This leads us to introduce a \textbf{novel measure, which incorporates both of these factors -- the Factual Robustness Score (FRS)} -- which is further detailed in \cref{sec:frs_main}.




\section{Related Work} 

\subsection{Factual Knowledge in Language Models}  

LLMs encode vast amounts of factual knowledge, often retrieved from internal representations rather than structured databases.
While models can recall these facts, their responses remain inconsistent, suffering from hallucinations and retrieval errors (\citet{Petroni2019Knowledge}, \citet{Roberts2020HowMuch}). A prominent research direction focuses on understanding and modifying how knowledge is encoded within LLMs: while \citet{geva2020transformer} and \citet{dai2021knowledge} identified specific structures within Transformers, such as key-value memory stores and knowledge neurons, \citet{de2021editing} and \citet{meng2022locating} explored techniques to directly manipulate the information stored within model weights. However, beyond the challenge of modifying knowledge, ensuring its stability and reliability remains an open problem. While \citet{zong2024comparisonqa} introduced ComparisonQA to evaluate factual robustness under controlled knowledge frequency and uncertainty, a broader body of work has framed robustness through the lens of adversarial attacks. Studies such as \citet{xu2024llmfoolitself}, \citet{bondarenko2024llm}, and \citet{howe2024exploring} have examined the vulnerability of LLMs to prompt-side perturbations, revealing inconsistencies in factual recall. Approaching the question through the prompt lens, \citet{mahaut2024factual} consider factual robustness in terms of semantically equivalent question phrasing. Contrary to the approaches above, \textbf{our work shifts the focus from external prompt modifications to the model-side of the generation process}, investigating the internal mechanisms that contribute to factual robustness.

\subsection{Temperature and Language Models} 

While seemingly distant from modern LLMs, \citet{ackley1985learning} established temperature as a key factor in shaping probability distributions, a principle still central to language model decoding today. Recent studies have debated whether temperature significantly impacts problem-solving ability. \citet{Renze24Sampling} found that varying temperature from 0.0 to 1.6 had no significant effect on accuracy, challenging the assumption that lower temperatures enhance reasoning. Beyond accuracy, temperature is often associated with creativity: \citet{peeperkorn2024temperature} showed that while higher temperatures increase variation, this does not necessarily translate into meaningful novelty, raising questions about its role.  

Recognizing the limitations of fixed-temperature sampling, researchers have developed adaptive temperature control strategies to balance quality and diversity. \citet{chang2023kl} introduced KL-Divergence Guided Temperature Sampling, which adjusts temperature dynamically based on token relevance, while \citet{zhang2024edt} proposed Entropy-based Dynamic Temperature (EDT) Sampling to optimize fluency and diversity. 

A novel perspective on temperature scaling comes from \citet{nakaishi2024critical}, who found that LLMs undergo critical phase transitions at certain temperature thresholds, exhibiting abrupt behavioral shifts akin to phase changes in physical systems. This suggests that temperature adjustments do more than refine probability distributions -- they can fundamentally reshape generative dynamics, making temperature a key factor in understanding LLM computational behavior.    

\paragraph{Our Contribution.}
Unlike existing performance-based methods that assess correctness at fixed temperatures, or studies in attack-based scenarios, we evaluate how robustly a fact is stored within a model by incorporating both entropy and ``breaking'' temperature -- the temperature at which a model switches to producing an incorrect answer. This allows for a more comprehensive assessment of factual stability, moving beyond surface-level correctness to quantify how knowledge withstands perturbations in sampling conditions. Our work provides new insights into the resilience of LLMs and highlights the need for robustness-focused evaluation metrics in future research.  

\section{Experiments and Results}

\subsection{Experimental Setup}
\subsubsection{Closed-Book Question Answering Setup}

Our study operates within a closed-book question answering (QA) setting, where models generate answers based solely on the input question, without access to external documents or retrieval mechanisms. This ensures that any correctly answered question reflects knowledge stored within the model's parameters.

\noindent\textbf{Datasets.} We use three widely studied QA datasets, each adjusted for the closed-book setting: (1)    SQuAD \citep{rajpurkar-etal-2016-squad}, a reading comprehension dataset primarily composed of Wikipedia-based questions; (2) TriviaQA \citep{2017arXivtriviaqa}, a collection of knowledge-intensive questions originally designed for open-domain QA; and (3) HotpotQA \citep{yang2018hotpotqa}, a multi-hop QA dataset that requires reasoning over multiple facts.

\noindent\textbf{Model Selection.} To ensure diversity in model architectures, scales, and training sources, we evaluate five different LLMs: i.e., GPT-4o-mini \citep{achiam2023gpt}, LLaMA-3.2-3B and LLaMA-3.1-8B \citep{touvron2023llama}, Qwen-2.5-3B and Qwen-2.5-14B \citep{yang2024qwen2}. The GPT models are accessed via the OpenAI API, while the LLaMA and Qwen models are loaded from Hugging Face, using their \texttt{instruct} fine-tuned versions.

\noindent\textbf{Controlling Response Length and Format.} To manage verbosity and enforce concise answers, we provide each model with two-shot exemplars -- examples of questions paired with expected, brief responses -- before prompting with the actual question. Additionally, we set \texttt{max\_new\_tokens=5}, limiting response length to a maximum of five tokens. This setup ensures that models produce concise, directly comparable outputs across datasets and temperature conditions, enabling a rigorous analysis of factual recall and response certainty.


\subsubsection{Selection of Correctly Answered Questions for Analysis}
To establish a controlled and reliable basis for our experiments, we begin by identifying $1000$ correctly answered questions per dataset and model. This selection process ensures that our analysis focuses on instances where the model demonstrably possesses the required knowledge.

\noindent\textbf{Baseline Selection Criteria.} To identify these correctly answered questions, we set the generation temperature $t$ to zero, enforcing deterministic output selection. This choice minimizes stochastic variability, ensuring that the most probable token sequence is always selected. Since we study model \textbf{robustness}, we are showing the effects of increasingly hard generation conditions. Hence, by first setting $t=0$, we establish a baseline representing the model's most ``ideal'' conditions for factual retrieval. Additionally, to simulate a well-calibrated model, we filter out instances where the model confidently produces incorrect answers. This ensures that our study focuses solely on high-confidence correct responses, allowing us to analyze the impact of increasing temperature without confounding effects from miscalibrated predictions.

\noindent\textbf{Inference and Selection Process.} At inference time, the model generates a single response per question. We evaluate the output using an exact match criterion -- if the generated answer is identical to the ground truth, the sample is retained as a correctly answered question. This guarantees that the model has the knowledge necessary to answer these questions correctly under optimal conditions. By enforcing these selection criteria, we create a robust experimental foundation that allows us to systematically study the effects of increasing temperature on response accuracy and certainty.




\begin{figure*}[!t]
     \centering
     \includegraphics[width=\linewidth]{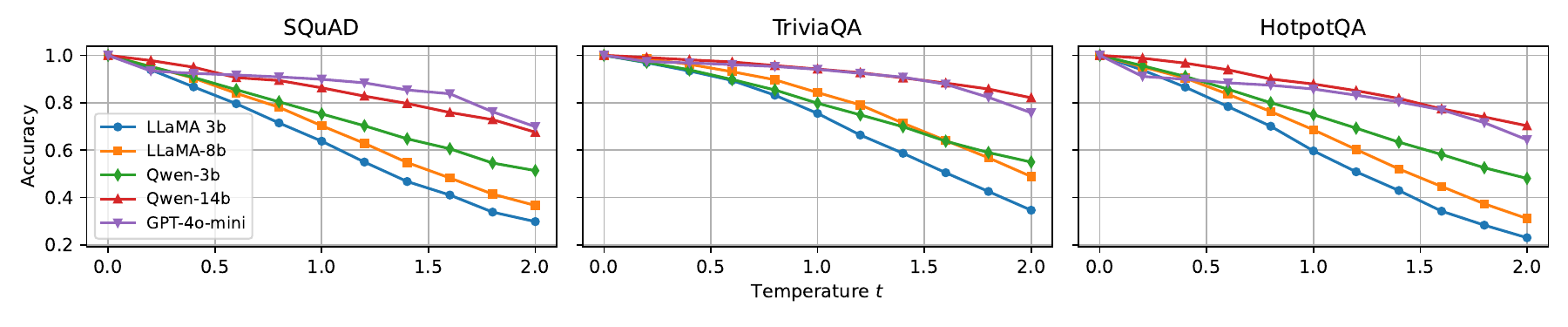}
   \caption{\textbf{Accuracy steadily degrades with increasing temperature levels.} We show that temperature is a direct factor in the difficulty of keeping the correct answer. Across all LLMs and datasets, accuracy decreases with increasing temperature, making temperature a direct factor in correct responses.}
   \label{fig:temp_vs_acc}
\end{figure*}

\subsection{Temperature Scaling}\label{sec:temperature_scaling}  

Temperature \( t \) is a key parameter in the generation process, controlling the level of randomness and semantic variability in model outputs. It theoretically ranges from \( 0 \) to $\infty$, although it is typically constrained between \( 0 \) and \( 2 \) \citep{Renze24Sampling}. Recall that lower temperatures lead to more deterministic outputs, while higher values introduce greater diversity by flattening the probability distribution.  To systematically analyze the impact of temperature on response accuracy and certainty, we apply temperature scaling to the previously identified correctly answered questions for each model and dataset. The procedure is as follows:  

\noindent\textbf{(1)} For each question, we progressively increase \( t \) and observe performance variations. 

\noindent\textbf{(2)}  Since generation is a probabilistic sampling process, we run each question 10 times per temperature setting to account for variability in model outputs.

\noindent\textbf{(3)} A response is deemed correct if it contains the ground truth answer, allowing for minor verbosity while maintaining correctness beyond exact match. Accuracy is then calculated as the proportion of correct responses across 10 trials per question.

\noindent We evaluate \( t \in \{0.2, \dots, 2.0\} \), with steps of \( +0.2 \). This range ensures a balanced exploration of temperature effects, spanning from near-deterministic outputs to moderately diverse sampling regimes. The following results illustrate how temperature scaling influences factual robustness across different models and datasets.

\paragraph{Effects of Temperature on Factual Accuracy and Certainty Levels.} As shown in \cref{fig:temp_vs_acc}, accuracy consistently decreases with increasing temperature across all models and datasets. This decline is a direct consequence of higher temperature values amplifying probabilistic sampling, allowing less probable tokens to be selected more frequently. 
Notably, smaller models such as LLaMA-3b and LLaMA-8b experience the steepest declines, losing over 60\% of their accuracy in some cases. In contrast, larger models like GPT-4o-mini and Qwen-14b demonstrate greater resilience, maintaining relatively higher accuracy at elevated temperatures. This trend suggests that model size and training scale influence robustness to temperature-induced uncertainty, with larger models better preserving factual consistency despite increased stochasticity in token selection. 

\begin{figure*}[!t]
    \centering
    \includegraphics[width=\textwidth]{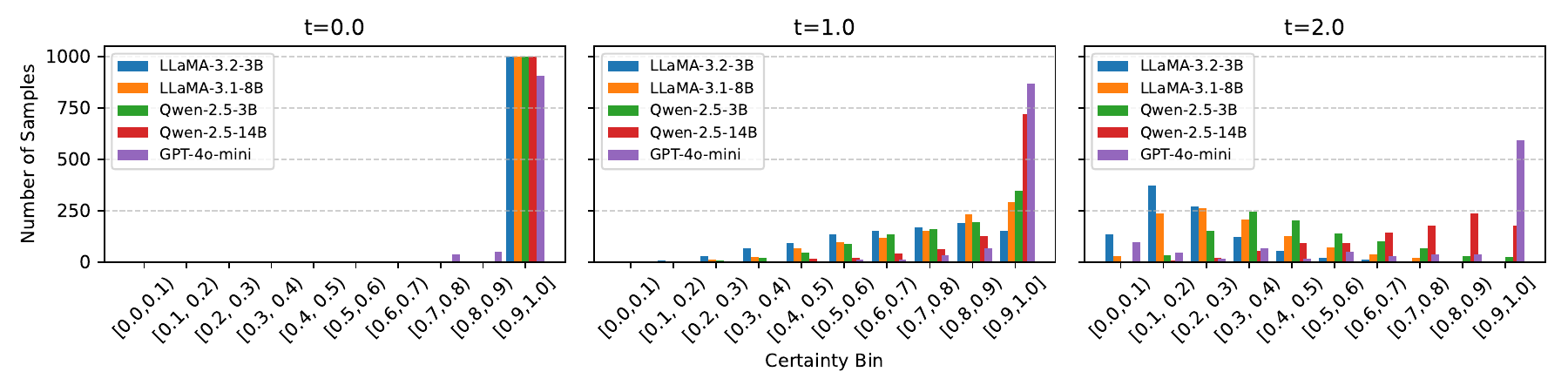}
    \caption{\textbf{Impact of temperature $\mathbf{t}$ on token probability distribution in TriviaQA}. As $t$ increases, the probability distribution flattens, reducing certainty in token selection and increasing the likelihood of generating lower-probability responses. This highlights how temperature directly influences response confidence and factual stability.}
    \label{fig:modified_temp_certainties}
\end{figure*}

Furthermore, we analyze the certainty levels, in terms of token probability, of generated responses at varying temperature settings (\cref{fig:modified_temp_certainties}). Here, we compute the average probability of generated responses and categorize them into certainty bins.\footnote{For example, the bin 0.7-0.8 includes all responses -- correct or incorrect -- whose average probability falls within the range $[0.7,0.8)$.} This visualization clearly demonstrates how temperature scaling flattens the probability distribution, increasing competition among tokens and making the correct answer less consistently selected. Interestingly, even as temperature increases, some responses remain correct, indicating that certain facts are inherently more robust to temperature-induced variability. This observation suggests a potential for defining a \textbf{temperature-based robustness metric}, where the level of $t$ at which a correct answer ``breaks'' could serve as a measure of factual stability within the model.



\subsection{Entropy of true distribution}\label{sec:entropy_true_distribution}

Besides temperature, we analyze the entropy levels of generated answers to understand how entropy interacts with temperature-induced distribution shifts without compromising factual accuracy. To standardize our entropy measurements, we use \(\log_{10}\) instead of \(\log_2\), ensuring that the maximum entropy value is fixed at 1. The theoretical upper bound of entropy for a given probability distribution is determined by \(\log(n)\), where \(n\) represents the number of possible outcomes -- in our case, the number of candidate tokens at a given position. Since we restrict the top token choices to 10\footnote{The restriction is necessary due to API limitations for GPT-4o-mini. In order to be consistent, we apply the same limit to all models.}, the maximum entropy value in our setup is \(\log_{10}(10) = 1\).

\paragraph{Entropy vs. Breaking Temperature.} We begin by analyzing the relationship between the baseline entropy of correct answers -- i.e., the entropy over correctly generated tokens at \( t=0 \) -- and the temperature at which these answers become incorrect due to token selection shifts. For each sample, we determine the first temperature level where the model’s accuracy drops below a predefined threshold of 0.5, hence we say that the model ``breaks''.\footnote{We define a significant degradation as an accuracy drop below 50\%.} Once the breaking temperature \( t_b \) is identified for each sample, we compute the entropy of the originally correct answer's token distribution at \( t=0 \). Entropy is calculated at each token position and then averaged across all tokens in the answer, yielding a single entropy score per response. 

\subsection{Combining Entropy and Temperature}\label{sec:entropy_vs_temp}
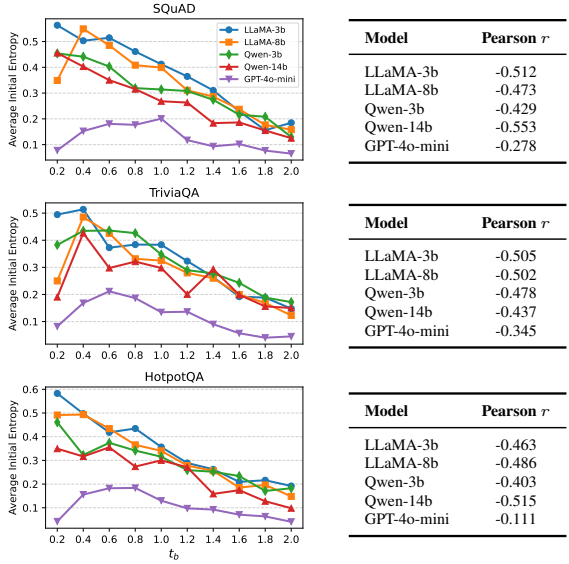
\begin{figure}[!t]
\centering
\begin{tikzpicture}
    \node[anchor=south west, inner sep=0] (image) at (0,0) {
        \includegraphics[width=\linewidth]{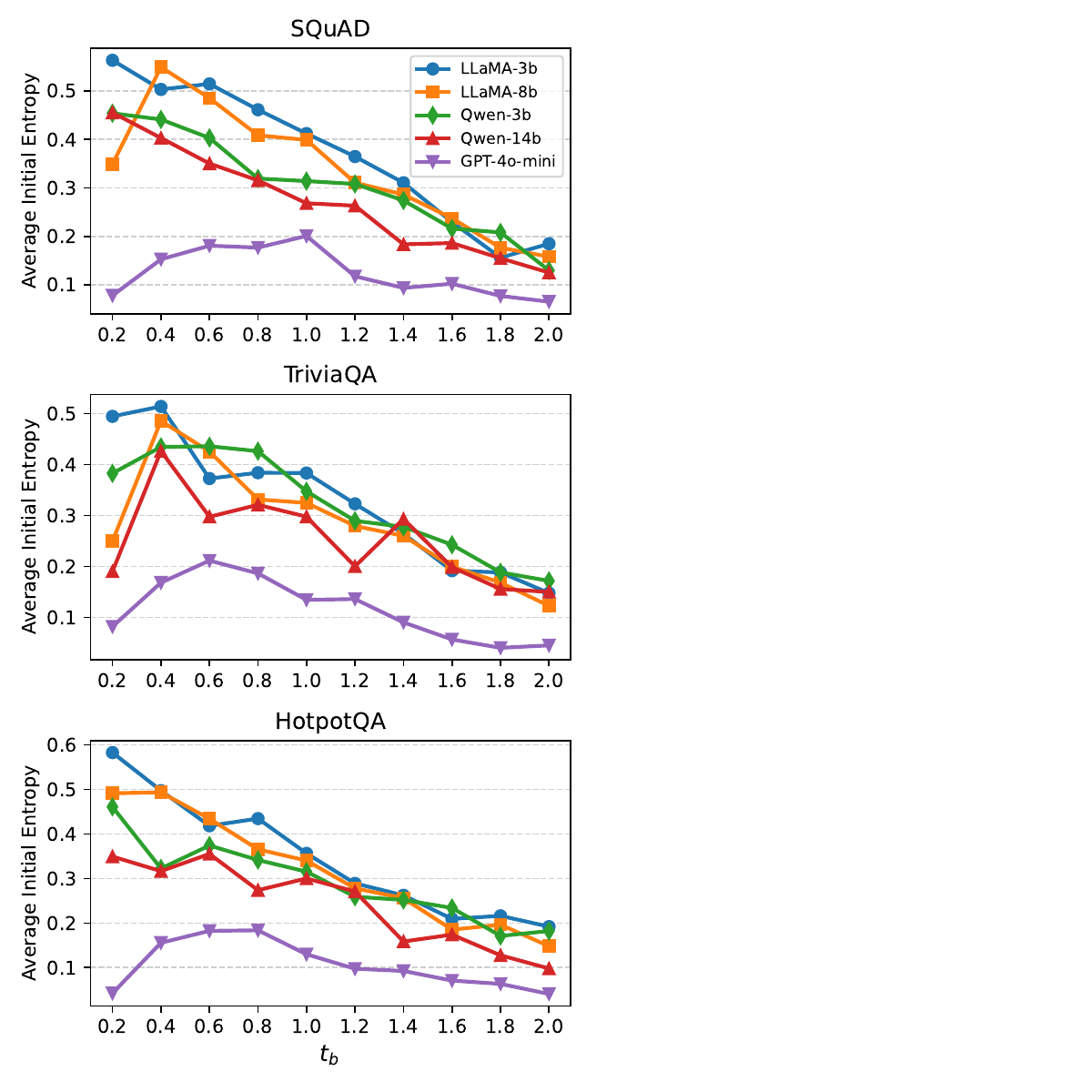}
    };

    \node[anchor=north west] at (4.5, 7.5) { 
        \tiny
        \begin{tabular}{lc}
        \toprule
        \textbf{Model} & \textbf{Pearson $r$} \\
        \midrule
        LLaMA-3b & -0.512 \\
        LLaMA-8b & -0.473 \\
        Qwen-3b & -0.429\\
        Qwen-14b & -0.553 \\
        GPT-4o-mini & -0.278 \\
        \bottomrule
        \end{tabular}
    };
    \node[anchor=north west] at (4.5, 5.05) {
    \tiny
    \begin{tabular}{lc}
            \toprule
            \textbf{Model} & \textbf{Pearson $r$} \\
            \midrule
            LLaMA-3b & -0.505 \\
            LLaMA-8b & -0.502 \\
            Qwen-3b & -0.478 \\
            Qwen-14b & -0.437 \\
            GPT-4o-mini & -0.345 \\
            \bottomrule
            \end{tabular}
    };
    \node[anchor=north west] at (4.5, 2.55) {
            \tiny
            \begin{tabular}{lc}
            \toprule
            \textbf{Model} & \textbf{Pearson $r$} \\
            \midrule
            LLaMA-3b & -0.463 \\
            LLaMA-8b & -0.486 \\
            Qwen-3b & -0.403 \\
            Qwen-14b & -0.515 \\
            GPT-4o-mini & -0.111 \\
            \bottomrule
            \end{tabular}
    };

\end{tikzpicture}
\caption{\textbf{Average entropy levels of originally correct
answers vs. the breaking temperature levels.} While we observe a downward trend, we show that there is only a weak correlation between the initial entropy of an answer and its breaking temperature.} 
\label{fig:entropy_vs_break_temp}
\end{figure}

After establishing the relevance of both entropy and temperature scaling to the concept of factual robustness, we investigate the correlation between the two factors in \cref{fig:entropy_vs_break_temp}. The plots depict the initial entropy levels in relation to the breaking temperature. While there is a downward trend to be observed, the two variables do not exclusively predict each other. We show this by computing the Pearson Correlation between the two variables of initial entropy and breaking temperature, and find only a slight negative correlation for most models ($-0.48$ on average), while GPT-4o-mini shows almost no correlation ($-0.24$ on average).

Since we show that these two concepts \textbf{do not express the same dimension of factual robustness} yet are both elementary to the robustness concept, we proceed to combine them into one solid Factual Robustness Score.

\section{How Factually Robust are LLMs?}
\label{sec:frs_main}
In contrast to the findings of \citet{Renze24Sampling}, who argue that temperature has no significant impact, our results in \cref{sec:temperature_scaling,sec:entropy_true_distribution} demonstrate that increasing temperature in a QA context reduces both accuracy and certainty, regardless of model size.  This sensitivity to temperature highlights the need for LLMs to maintain factual reliability under varying generation conditions. 

\textbf{Our goal is to provide a quantitative measure of factual robustness in LLMs}. To this end, we introduce the Factual Robustness Score (FRS), which captures the stability of factual knowledge and its resilience to distributional shifts.  From our perspective, a fact's robustness depends on two key factors: \textbf{(1)} the model's initial uncertainty (entropy) when generating the fact at \( t=0 \), and \textbf{(2)} its resistance to perturbations as temperature increases. By integrating both aspects, we propose FRS as a single, comprehensive score that quantifies how well an LLM retains factual knowledge under varying temperature sampling conditions.

\subsection{Factual Robustness Score (FRS)}\label{sec:frs}

\textbf{Initial Confidence.} We quantify the model's initial certainty by $(1-H)^d$ where $H \in [0,1]$ is the entropy -- scaled so that \(H=1\) means high uncertainty and \(H=0\) means high confidence -- and $d \geq 1$ tunes how strictly we penalize uncertainty -- i.e., larger $d$ penalizes even moderate $H$ more severely. \\

\noindent\textbf{Temperature Resilience.} Let \(\,t_b \ge 0\,\) be the smallest temperature at which the model's accuracy drops below a chosen threshold (e.g., 50\%). We refer to it as the \textit{breaking temperature}. If \(\,t_b\) is large, the model remains correct under stronger sampling perturbations and should receive a higher score. To capture this, we include the factor \(\,(t_b+1).\)\\

\noindent\textbf{Entropy-Based Penalty.} Even if the model can withstand high temperatures, large \(H\) indicates the model is essentially ``guessing.'' Hence we subtract a penalty proportional to \(H\). For scale consistency, we divide by \((t_b+1)\) to reduce the penalty when \(t_b\) is large: i.e., $\frac{H}{t_b+1}$.\\

\noindent\textbf{Constructing the Final Formula}. We combine the ``reward'' and the ``penalty'' into a single function $f: \mathbb{R}_0^1 \times \mathbb{R}_+ \times \mathbb{R}_+ \to \mathbb{R}$:
\begin{equation}\label{eq:frs}
f(H,d,t_b)
\;=\;
(1 - H)^d\cdot(t_b + 1)-\frac{H}{t_b + 1}.
\end{equation}

\begin{figure*}[t!]
    \centering
    \includegraphics[width=\linewidth]{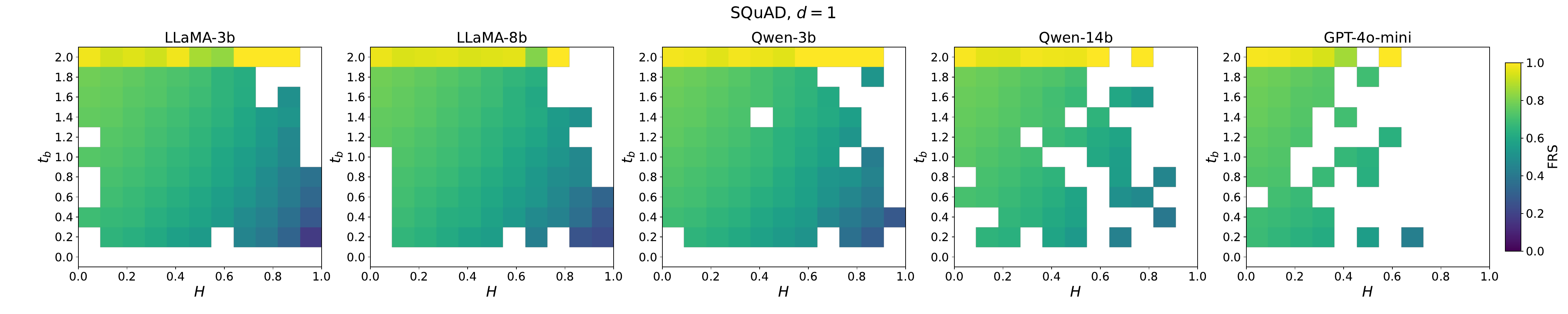}
    \caption{FRS ($d=1$) over all models, on SQuAD. Although FRS equal to 1 is theoretically achievable when $t_b\to\infty$, in practice, we set it to 1 for all samples where models did not break with $t_b \leq 2$. Hence, the yellow data points.}
    \label{fig:frs_2d}
\end{figure*}


\noindent Since $f(H,d,t_b)\in [-1,\infty)$ -- see~\cref{app:bounds} -- we slightly modify~\cref{eq:frs} to have a more interpretable score of factual robustness by balancing initial knowledge confidence (entropy) with resilience to temperature perturbations as follows:
\begin{equation}\label{eq:scaled_frs}
    f(H,d,t_b)_0^1 =  \frac{f(H,d,t_b)+1}{f(H,d,t_b)+2}.
\end{equation}%
We refer the reader to~\cref{app:transformation_frs} for a step-by-step derivation of~\cref{eq:scaled_frs}.

\paragraph{Intuition of FRS.}
We further visualize the intuition behind the function in~\cref{eq:frs} and how the entropy is a weighting factor over the breaking temperature (see \cref{app:c_intuition}). Recall that a model retains a fact if it obtains at least $50\%$ accuracy in answering over 10 trials; otherwise, it breaks. When \(H \approx 0\) (high confidence) and \(t_b\) is large (breaks late), \(f\) becomes large. When \(H \approx 1\) (no confidence), \((1-H)^d \approx 0\) and \(\,H/(t_b+1)\approx 1/(t_b+1),\) so \(f\) is near zero or negative. 
Increasing \(d\) tightens the penalty on any uncertainty \(H>0\), reflecting stricter requirements for robust factual knowledge. This derivation shows how \(\,f(H,d,t_b)\) cleanly fuses \textbf{how confident a model is} in a fact with \textbf{how long it withstands} temperature perturbations, thus producing a single measure of factual robustness.

\begin{table}[!t]
    \centering
    \caption{Average FRS with $d \in \{1, 2, 5, 10, 50\}$ across all datasets. Generally larger \textbf{models are more robust than smaller ones}. $^\star$ indicates the most robust model overall.}
    \label{tab:llm_comparison}
    \resizebox{\linewidth}{!}{%
\begin{tabular}{lccccc}
        \toprule
    $d$ & LLaMA-3b & LLaMA-8b & Qwen-3b & Qwen-14b  $^\star$ & GPT-4o-mini \\
    \midrule
    $1$ & $0.761$  & $0.812$  & $0.855$  & $\mathbf{0.935}$ & $0.923$ \\
    $2$ & $0.727$  & $0.787$  & $0.836$ & $\mathbf{0.928}$ & $0.919$ \\
    $5$ & $0.669$  & $0.741$  & $0.799$ & $\mathbf{0.913}$ & $0.910$ \\
    $10$ & $0.629$  & $0.706$  & $0.771$  & $0.898$ & $\mathbf{0.900}$  \\
    $50$ & $0.587$  & $0.663$  & $0.740$  & $\mathbf{0.878}$ & $0.875$ \\
    \bottomrule
\end{tabular}
    }
\end{table}

\subsection{Model Family vs. Size in Robustness} \cref{tab:llm_comparison} compares FRS across different model sizes and families using various values of \( d \in \{1,2,5,10,50\} \), controlling the influence of the entropy. As expected, larger models generally tend to be more robust than their smaller counterparts. However, a particularly interesting finding is that \textbf{model size alone is not the sole predictor of robustness across different model families}. While larger models within the same family (e.g., LLaMA-3b vs. LLaMA-8b) tend to retain facts more reliably, comparing across families (e.g., LLaMA vs. Qwen) introduces additional factors beyond just model size. Even though LLaMA-8b exhibits a more stable entropy trend in relation to \( t_b \) (see \cref{fig:entropy_vs_break_temp}), the smaller Qwen-3b model actually achieves a higher average FRS. Interesting in this context is also GPT-4o-mini, which, being an 8 billion parameters model, has similarly high robustness as the leader, Qwen-14b. Hence, we argue that different models store and retrieve factual knowledge in distinct ways. While parameter size plays a central role in a model's factual robustness, factors like model architecture and training data may also be defining factors.

We also show how the FRS gets hampered with an increasing $d$ based on the initial entropy. Notice how the LLaMA models suffer the highest drop in FRS ($-0.174$ for 3b, and $-0.149$ for 8b), meaning that they were uncertain, on average, about the initial factual knowledge, confirmed in~\cref{fig:temp_vs_acc} where LLaMA has the highest accuracy drop across the board. 
To explore this further, \cref{fig:frs_2d} illustrates FRS with \( d = 1 \) across all models on the SQuAD dataset. Notably, Qwen-14b achieves higher FRS scores even when its initial entropy is lower than that of GPT-4o-mini, whereas the latter's scores are more concentrated toward the lower entropy range. Additionally, GPT-4o-mini exhibits a more skewed probability density distribution, with $\text{FRS}\rightarrow 1$, suggesting that while it may be highly confident in certain cases, its robustness is not consistently superior across all factual scenarios (see \cref{app:g_probdens}). These findings highlight the need for a deeper examination into how architectures, training methodologies, and knowledge retention influence factual robustness.

\subsection{Most and Least Robust Facts}\label{sec:most_vs_least_robust_facts}
\begin{figure}[h]
    \includegraphics[width=\columnwidth]{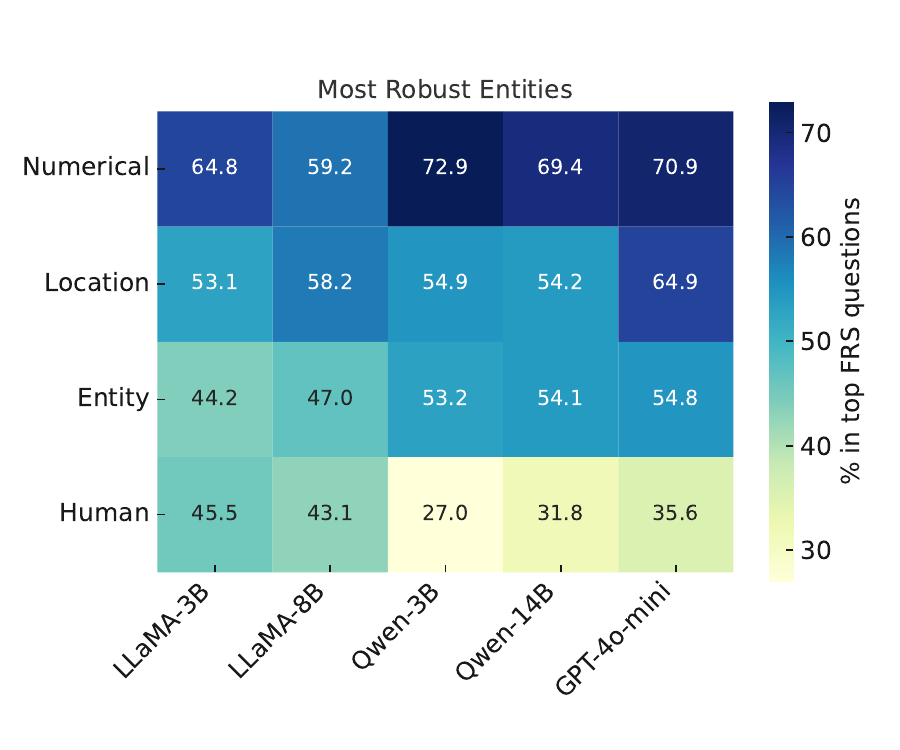}
    \caption{\textbf{Numerical facts are most robust across all models.} Grouping the top most robust answers into question categories across datasets, we show the dominance in \% of the categories for each model. Answers to numerical or location-based questions are most robust, while questions about names (here: \textit{Human}) lead to least robust answers.}
    \label{fig:frs_entities_heatmap}
\end{figure}

We further analyze the specific question types that are the most and least robust within the models. For this, we choose to fine-tune a lightweight classifier, DistilBERT \citep{sanh2019distilbert}, on the TREC question type dataset \citep{li2002learning} to classify the categories \textit{Numerical, Location, Entity} and \textit{Human}, for which we reach 96,6\% accuracy on the test set.\footnote{We report details about the dataset and the training procedure in \cref{app:h_questions}.} We then collect the top 1000 highest and lowest FRS score samples across all datasets, for each model. For both the top and bottom samples, we compute the proportion of each question type relative to the total number of samples of that type. \cref{fig:frs_entities_heatmap} shows the ratio of the most robust question types in percentage for each model. For example, for LLaMA-3B, 64.8\% of numerical type questions are among the most robust (while $100\%-64.8\%=35.2\%$ are part of the least robust).
\textbf{Answers associated with numerical and location related questions tend to be the most robust across models}, reaching up to 72.9\% and 64.9\%, respectively. The least robust answers belong to the \textit{Human} entity type, which means questions asking about the name of a person or a group. We show that \textbf{all models recall names in a less robust way}, meaning that those are most prone to be produced with high entropy and/or are shifting to a wrong answer at low temperature levels. 
We hypothesize that this is due to the nature of expected answers: for example, questions of the numerical type most often call for a single digit as answer, e.g., a year. Here, the LLM has less individual tokens to produce, and hence less chance of introducing an error, than for example with the name of a person, which is more lengthy. It is furthermore interesting to note the difference between model sizes: while the bigger models perform highest on numerical type questions, they underperform smaller models like LLaMA-3B on human type questions.

We want to emphasize that this study is not about certain LLMs performing better or worse on answering specific question types, but \textbf{the robustness of their answers}, which, according to our study, does not always directly correspond to model size and capabilities.


\section{Conclusion}

This study provides a comprehensive analysis of factual robustness in LLMs and introduces the Factual Robustness Score (FRS) as a novel metric for evaluating factual stability under varying conditions. Unlike traditional accuracy-based assessments, FRS captures both {internal generation uncertainty} (entropy) and {sensitivity to temperature}, offering a more nuanced measure of robustness. 

Our findings show that {increasing temperature reduces factual accuracy}, with models that are more confident in their initial answers maintaining correctness longer. However, since entropy alone does not fully predict breaking temperature $t_b$, FRS integrates both factors for a more complete evaluation. While {larger models generally exhibit greater robustness}, architectural and training differences mean that some smaller models outperform their larger counterparts. Additionally, {factual robustness varies by knowledge type}, with numerical facts proving more stable than others. In a practical sense, FRS  holds potential for enhancing a model’s factual accuracy through continued pre-training: by evaluating a model’s FRS, one can identify areas of weakness (specific facts where the model underperforms) and use these as targeted data for further training. Beyond introducing FRS, this study highlights limitations in current evaluation methodologies and demonstrates that factual robustness is influenced by both model properties and the nature of the knowledge itself. Future work could explore (1) finding the optimal breaking temperature for each model based on the initial entropy, (2) higher-temperature effects, and (3) strategies for enhancing factual retention to improve the reliability of LLMs in real-world applications.

\section*{Limitations}

One key limitation of FRS is the need to evaluate model responses across multiple temperature levels to determine the breaking point \( t_b \), making the process computationally intensive. Developing more efficient estimation methods to approximate robustness without exhaustive sampling would enhance its practicality. Additionally, we acknowledge that our analysis is constrained by an upper bound of \( t_b = 2.0 \), meaning we cannot determine at which \( t > 2.0 \) certain answers would break, leaving some robustness thresholds uncertain - similarly to so far uninvestigated temperatures in-between our chosen values of $t$. Lastly, our study focuses on question answering in a controlled, closed-book setting, whereas real-world applications often involve external knowledge sources, such as retrieval-augmented generation (RAG) systems, which may further impact factual stability.

\section*{Ethics Statement}
This work aims to analyze the behavior and reliability of large language models, contributing to a deeper understanding of their factual robustness. While our findings have potential societal implications, we do not identify any immediate ethical concerns that require specific attention. We encourage further discussions on the broader impact of LLM evaluation methodologies.

\bibliography{custom}

\appendix

\section{Temperature Effect on Probability Distribution}

\begin{figure*}
    \centering
    \includegraphics[width=\linewidth]{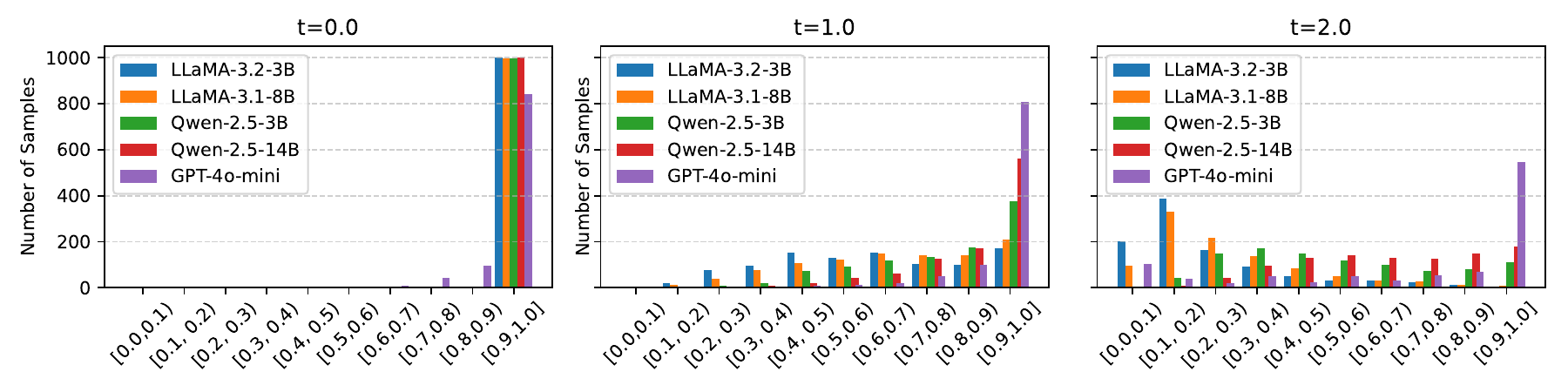}
    \caption{Impact of $t$ on token probability distribution: SQuAD.}
    \label{fig:temp_probs_squad}
\end{figure*}

\begin{figure*}
    \centering
    \includegraphics[width=\linewidth]{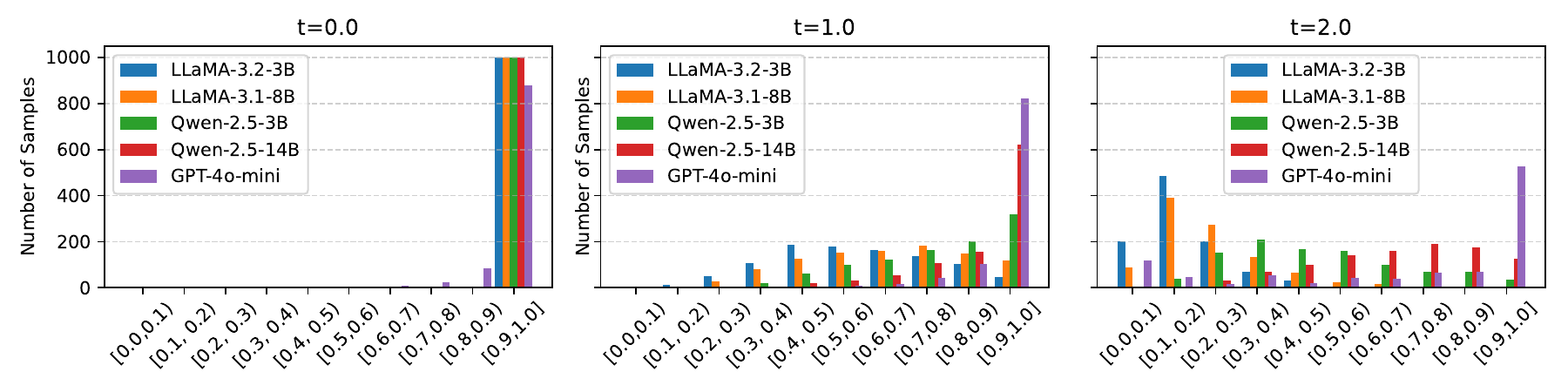}
    \caption{Impact of $t$ on token probability distribution: HotpotQA.}
    \label{fig:temp_probs_hotpot}
\end{figure*}

Identically to \cref{fig:modified_temp_certainties}, we provide visualizations of the token probability distributions under different temperatures for SQuAD and HotpotQA (see \cref{fig:temp_probs_squad,fig:temp_probs_hotpot}).

\section{Derivations on FRS}

For simplicity of the formulas, we suppose that $d = 1$, thus we omit it from the equations. However, regardless of the choice of $d$, the same bounds hold.

\subsection{Bounds of~\cref{eq:frs}}\label{app:bounds}
Given that $H\in[0,1]$ and $t_b \geq 0$, we first find the minimum and maximum values of~\cref{eq:frs}.

\subsubsection*{Case 1: $t_b = 0$}
\[
f(H,0) = (1 - H)(0 + 1) - H \cdot \frac{1}{0 + 1}
\]
\[
= (1 - H) - H = 1 - 2H
\]
When \( H = 0 \), we get \( f(0,0) = 1 \). When \( H = 1 \), we get \( f(1,0) = -1 \). Since \( f(H,0) \) is linear in \( H \), the range at \( t_b = 0 \) is $[-1,1]$.

\subsubsection*{Case 2: \( t_b \to \infty \)}

If $H<1$, we have $(1-H)\cdot(t_b + 1) \to \infty$ and $\frac{H}{t_b+1} \to 0$. Thus, when $H=0$, then $f(0,t_b) \to \infty$, and, when $H=1$, then $f(1,t_b) \to 0$.\\

\noindent Summarizing:
\begin{itemize}[noitemsep,topsep=0pt]
    \item At \( t_b = 0 \), the function is in the range \( [-1,1] \).
    \item As \( t_b \to \infty \), the function is in the range \( (0, \infty) \).
   \item Since \( f(H, t_b) \) is continuous and monotonic in \( t_b \), the full range of the function is $[-1,\infty)$.
\end{itemize}
\subsection{From~\cref{eq:frs} to~\cref{eq:scaled_frs}}\label{app:transformation_frs}

To scale the function
\begin{equation*}
    f(H, d, t_b) = (1 - H)^d \cdot (t_b + 1) - \frac{H}{t_b + 1}
\end{equation*}%
into the range \([0,1]\), we use min-max normalization:
\begin{equation}
    f_{0}^1(H, d, t_b) = \frac{f(H, d, t_b) - f_{\min}}{f_{\max} - f_{\min}},
\end{equation}%
where $f_{\min} = -1$ and $f_{\max} \to \infty$. Thus, we define:
\begin{equation}
\begin{aligned}
            f_0^1(H, d, t_b) = \frac{(1 - H)^d \cdot (t_b + 1) - \frac{H}{t_b + 1} + 1}{(1 - H)^d \cdot (t_b + 1) - \frac{H}{t_b + 1} + 2}\\
        = \frac{f(H,d,t_b)+1}{f(H,d,t_b) + 2}.
\end{aligned}
\end{equation}%

\section{Visual Intuition of FRS} \label{app:c_intuition}
\begin{figure}[!h]
    \centering
    \begin{minipage}{0.49\linewidth}
     \centering
     \includegraphics[width=\linewidth]{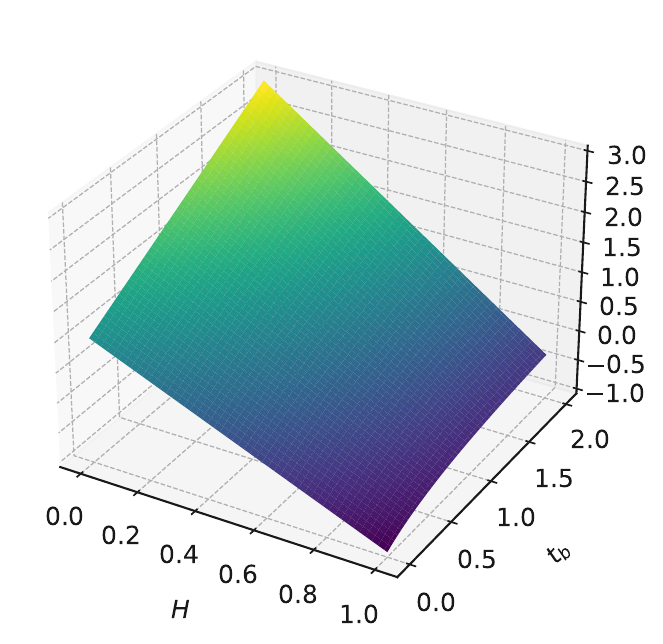}
   \end{minipage}\hfill
   \begin{minipage}{0.49\linewidth}
     \centering
     \includegraphics[width=\linewidth]{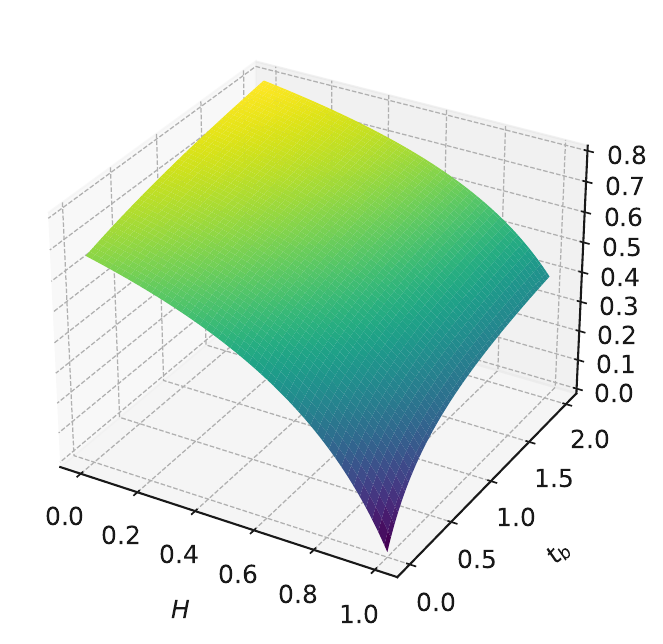}
   \end{minipage}\hfill
    \caption{3D plot of $f(H,1,t_b)$ and $f_0^1(H,1,t_b)$ over $H\in[0,1]$ and $t_b \in [0,2]$. The figure visualizes~\cref{eq:frs} (left) and~\cref{eq:scaled_frs} (right) where $H$ represents a weighting factor between two competing terms, and $t_b$ is the breaking temperature. The surface plot highlights how \textbf{FRS varies as a function of both parameters}.}
    \label{fig:frs}
\end{figure}

\begin{figure*}[!t]
    \centering
    \begin{minipage}{0.2\linewidth}
     \centering
     \includegraphics[width=\linewidth]{images/frs_scaled.pdf}
   \end{minipage}\hfill
   \begin{minipage}{0.2\linewidth}
     \centering
     \includegraphics[width=\linewidth]{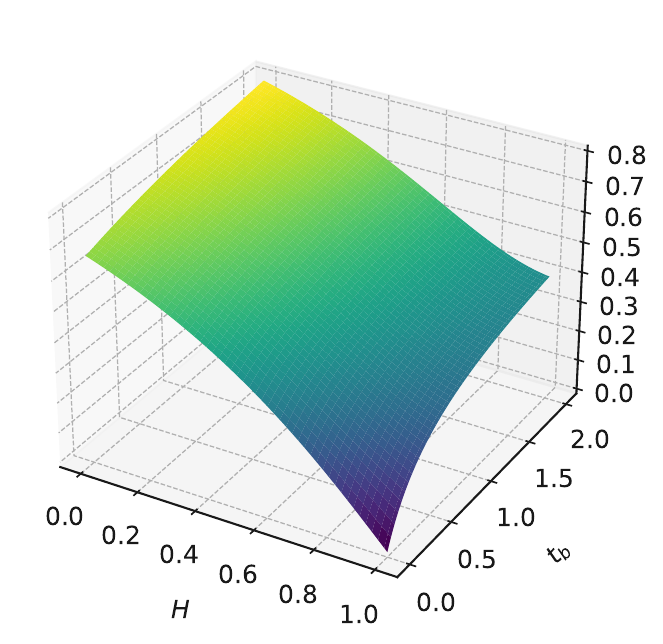}
   \end{minipage}\hfill
   \begin{minipage}{0.2\linewidth}
     \centering
     \includegraphics[width=\linewidth]{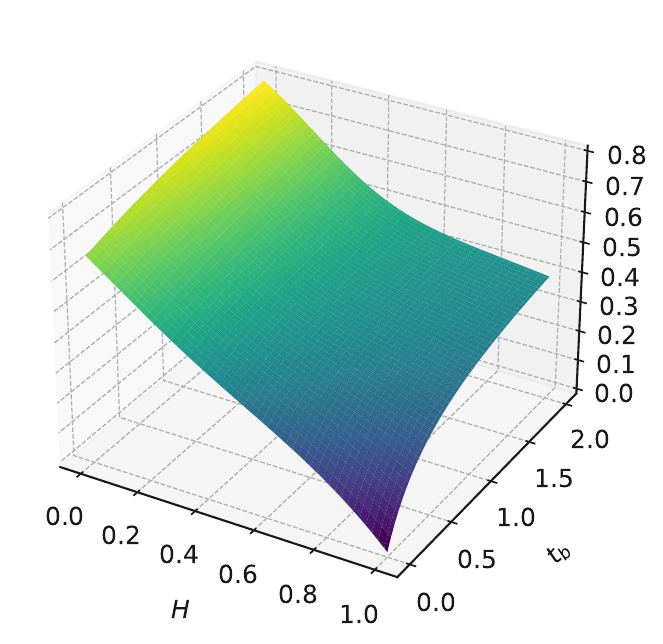}
   \end{minipage}\hfill 
   \begin{minipage}{0.2\linewidth}
     \centering
     \includegraphics[width=\linewidth]{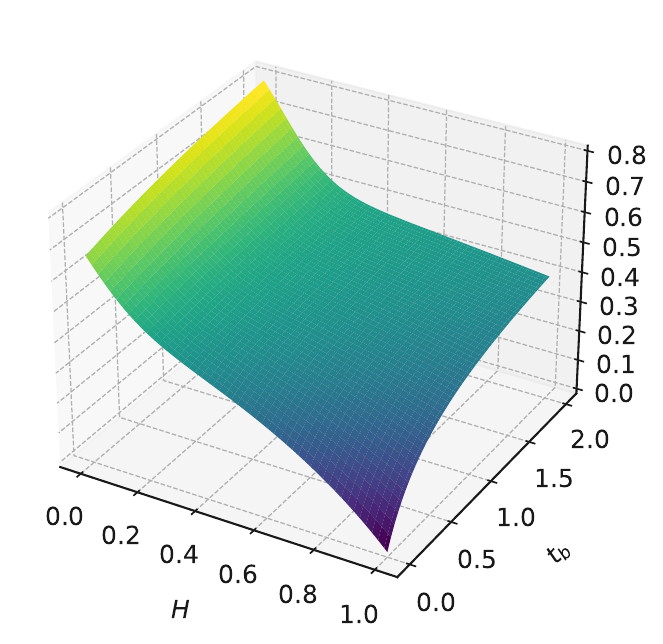}
   \end{minipage}\hfill
   \begin{minipage}{0.2\linewidth}
     \centering
     \includegraphics[width=\linewidth]{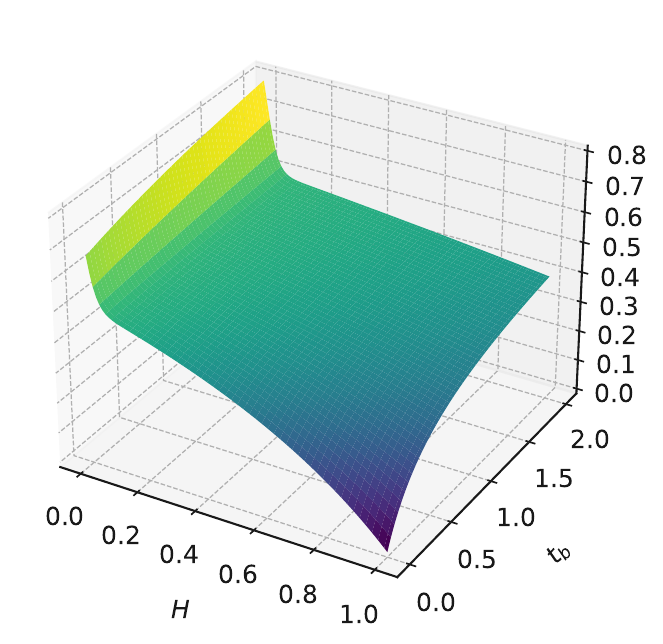}
   \end{minipage}\hfill
    \caption{3D plots of $f_0^1(H,d,t_b)$ over $H\in[0,1]$, $t_b \in [0,2]$, and $d \in \{1,2,5,10,50\}$ with $d$ varying from the left-most to the right-most subplot.}
    \label{fig:frs_with_varying_d}
\end{figure*}

As described in~\cref{sec:frs}, here we provide a visual intuition of the FRS function.~\cref{fig:frs} illustrates $f(H,1,t_b)$ and $f_0^1(H,1,t_b)$ by ranging the breaking temperature $t_b \in [0,2]$. Note how~\cref{eq:frs} -- see left plot -- is more aggressive (gives a lower FRS) with $t_b\to2$, while~\cref{eq:scaled_frs} has a smoother transition to higher FRS. Additionally, we show how $d$ reshapes the FRS function by giving more importance to the initial entropy $H$ -- see~\cref{fig:frs_with_varying_d}.

\section{Effect of \( d \) in FRS and its Role in Factual Robustness of LLMs}\label{app:d_effect}
The exponent $d$ of~\cref{eq:frs} plays a crucial role in adjusting the sensitivity of the factual robustness measure to entropy $H$. The term \( (1 - H)^d \) introduces a nonlinear scaling effect on the first component \( (t_b + 1) \). The impact of \( d \) is analyzed below and illustrated in~\cref{fig:frs_with_varying_d}.

\subsection*{Case 1: \( d > 1 \) (Stronger Penalization for High Entropy)}
\begin{itemize}[noitemsep,topsep=0pt]
    \item When \( d \) is large, even moderate values of \( H \) significantly reduce \( (1 - H)^d \), making the first term in \( f(H,d,t_b) \) much smaller.
    \item This causes the function to strongly penalize models with high entropy.
    \item This is desirable because high entropy corresponds to greater uncertainty in the model’s responses, making them less trustworthy.
\end{itemize}

\subsection*{Case 2: \( 0 < d < 1 \) (Smoother Decay of Robustness with Entropy)}
\begin{itemize}[noitemsep,topsep=0pt]
    \item If \( d \) is small, the decay of \( (1 - H)^d \) is much less aggressive.
    \item Even models with moderate entropy will not be penalized as strongly.
    \item This allows models with some uncertainty to still contribute positively to factual robustness.
\end{itemize}

\subsection*{Case 3: \( d = 1 \) (Linear Dependence on Entropy)}
\begin{itemize}[noitemsep,topsep=0pt]
    \item The function simplifies to:
    \begin{equation}
        f(H, 1, t_b) = (1 - H) (t_b + 1) - \frac{H}{t_b+1}.
    \end{equation}
    \item The effect of \( H \) on robustness is purely linear, with no additional weighting applied to lower-entropy or higher-entropy cases.
\end{itemize}

\subsection{Why is \( d \) Useful for Evaluating Factual Robustness?}
Introducing \( d \) allows for greater control over the sensitivity of factual robustness to entropy, making it particularly useful when evaluating the reliability of LLMs under adversarial conditions. When \( d > 1 \), models producing uncertain responses (\( H \to 1 \)) contribute less to factual robustness, which aligns with the intuition that uncertain responses are inherently less trustworthy. This ensures that models with higher entropy are penalized more aggressively, reflecting their reduced reliability.

Additionally, the robustness score should scale appropriately with \( t_b \), the breaking temperature at which the model begins to fail. Since the first term in \( f(H, d, t_b) \) is multiplied by \( (t_b + 1) \), models that can withstand higher adversarial pressures before failing naturally receive a higher robustness score. This scaling property ensures that models capable of maintaining factual accuracy under increasing difficulty are recognized as more robust.

Another key advantage of introducing \( d \) is that it provides a tunable sensitivity to entropy, allowing for customized evaluation criteria. A high value of \( d \) enforces stricter robustness criteria, meaning that even moderate uncertainty is penalized heavily. This is particularly useful in applications where high confidence is required, such as medical AI or legal text generation. Conversely, a lower value of \( d \) results in a more lenient evaluation, permitting models with some uncertainty to still be considered reasonably robust. This flexibility makes the function adaptable to various contexts, balancing the trade-off between strictness and tolerance in factual robustness assessment.

In general,  \( d \) acts as a tuning factor that determines how much we penalize uncertainty when assessing the robustness of an LLM. This flexibility is crucial when evaluating models under different conditions, such as \textbf{(a)} \textit{high-stakes applications} (e.g., medical AI) where high certainty is required, favoring a high $d$, and \textbf{(b)} \textit{general NLP tasks} where moderate uncertainty might be acceptable, allowing for a lower $d$.

\section{Time Complexity Analysis of FRS}

To compute the FRS, we consider the following steps:
\begin{enumerate}[noitemsep,topsep=0pt]
    \item Generating an answer using a transformer-based architecture.
    \item Finding the breaking temperature \( t_b \).
    \item Computing entropy \( H \).
    \item Computing the FRS function itself.
\end{enumerate}

\subsection{Feeding the Question to the LLM and Generating an Answer}
Modern autoregressive LLMs generate text one token at a time. If the model generates an answer of length \( L \), it requires \( \mathcal{O}(L) \) forward passes. Each forward pass involves a Transformer forward computation, which has complexity:
\begin{equation}
    \mathcal{O}(L \cdot D^2),
\end{equation}%
where $L$ is the number of generated tokens, and $D$ is the model's hidden size which scales with the number of parameters. Thus, the time complexity for generating one answer is $\mathcal{O}(L\cdot D^2)$.

\subsection{Finding the Breaking Temperature}
The breaking temperature \( t_b \) is the point where the model’s accuracy falls below a certain threshold -- in our case \( 50\% \). Determining \( t_b \) requires multiple evaluations.

\paragraph{Incremental Search:}
\begin{itemize}[noitemsep,topsep=0pt]
    \item We test up to \( \mathcal{O}(T) \) different values for \( t_b \).
    \item Each requires \( k \) model runs.
    \item Since each model run has \( \mathcal{O}(L \cdot D^2) \) complexity, the total cost is:
\end{itemize}%
\begin{equation}
    \mathcal{O}(T \cdot k \cdot L \cdot D^2)
\end{equation}

\paragraph{Binary Search (Optimized Approach):}
\begin{itemize}[noitemsep,topsep=0pt]
    \item A structured search can reduce the number of evaluations to \( \mathcal{O}(\log T) \).
    \item Each evaluation still requires \( k \) model runs.
    \item This reduces the complexity to:
\end{itemize}%
\begin{equation}
    \mathcal{O}(k \log T \cdot L \cdot D^2)
\end{equation}

\subsection{Computing Entropy}
The entropy of a generated sequence is computed as:
\[
H = -\sum_{i} P(x_i) \log P(x_i)
\]
where $P(x_i)$ are the token probabilities output by the LLM, and for each of the $L$ tokens, the model produces a probability distribution over a vocabulary $V$. Hence, retrieving token probabilities costs \( \mathcal{O}(V L) \) since each token's probability distribution has \( V \) elements, and computing the overall entropy has a complexity of $\mathcal{O}(VL)$.

\subsection{Computing the FRS Function}
The function
\[
f(H, d, t_b) = (1 - H)^d \cdot (t_b + 1) - \frac{H}{t_b + 1}
\]
requires basic arithmetic and exponentiation, which are computed in $\mathcal{O}(1)$.

\subsection{Total Time Complexity Per Question}
Summing up all contributions:

\paragraph{Using Binary Search (Optimized):}
\[
\mathcal{O}(k \log T \cdot L \cdot D^2 + V L)
\]

\paragraph{Using Incremental Search (Naïve Approach):}
\[
\mathcal{O}(T \cdot k \cdot L \cdot D^2 + V L)
\]

\subsection*{Takeaway}
\begin{itemize}
    \item LLM inference dominates the complexity, as it requires \( \mathcal{O}(L \cdot D^2) \) per model run.
    \item Finding \( t_b \) is the most expensive step, since it requires multiple model evaluations.
    \item Entropy computation \( \mathcal{O}(V L) \) is relatively small compared to LLM inference.
    \item Binary search significantly reduces complexity from \( \mathcal{O}(T \cdot k) \) to \( \mathcal{O}(k \log T) \).
\end{itemize}

\section{Extended FRS Results}
\begin{figure*}[htp]
\centering
\begin{minipage}{\linewidth}
     \centering
     \includegraphics[width=\linewidth]{images/frs_2d/frs_2d_d1_squad.pdf}
   \end{minipage}\hfill
\begin{minipage}{\linewidth}
     \centering
     \includegraphics[width=\linewidth]{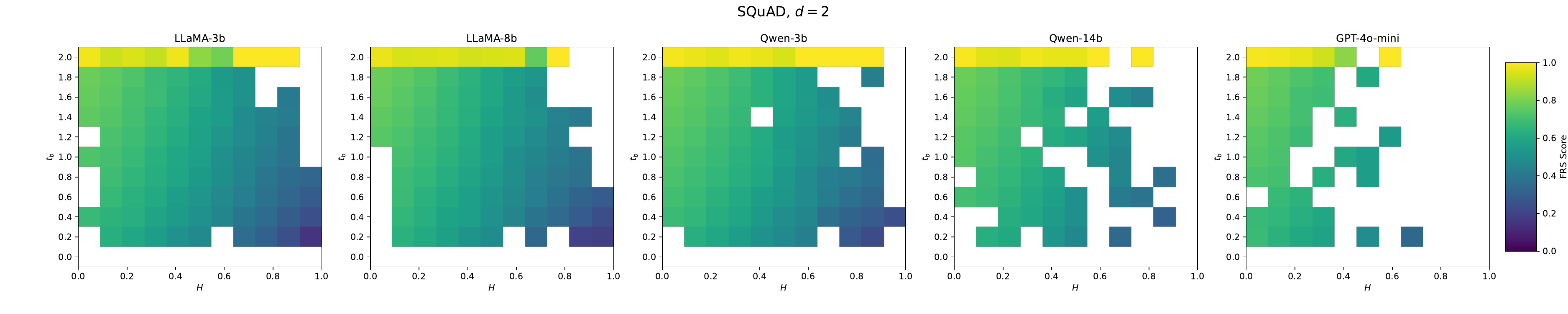}
   \end{minipage}\hfill
\begin{minipage}{\linewidth}
     \centering
     \includegraphics[width=\linewidth]{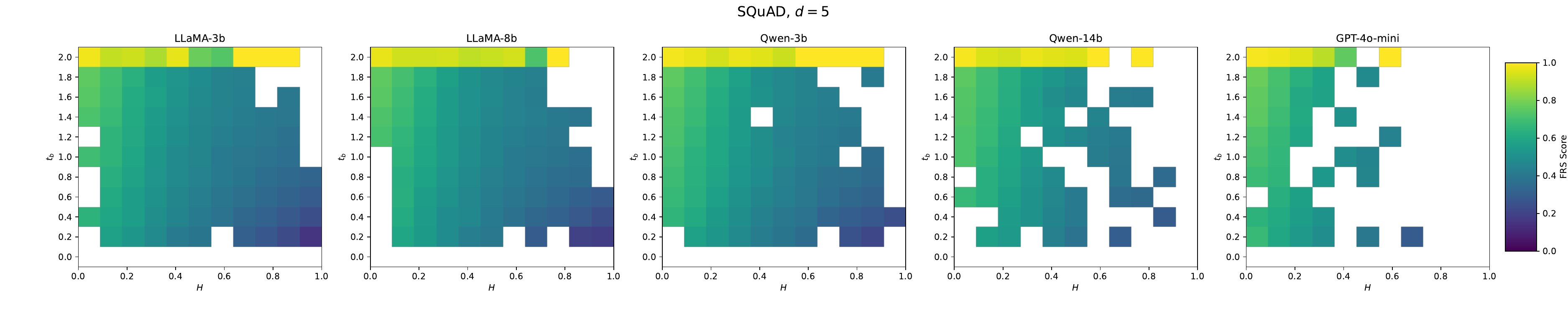}
   \end{minipage}\hfill
\begin{minipage}{\linewidth}
     \centering
     \includegraphics[width=\linewidth]{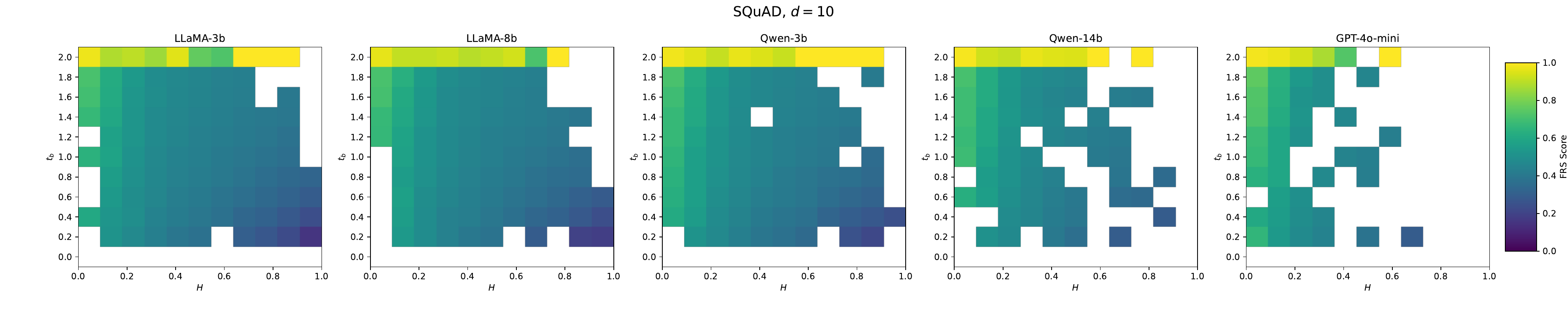}
   \end{minipage}\hfill
\begin{minipage}{\linewidth}
     \centering
     \includegraphics[width=\linewidth]{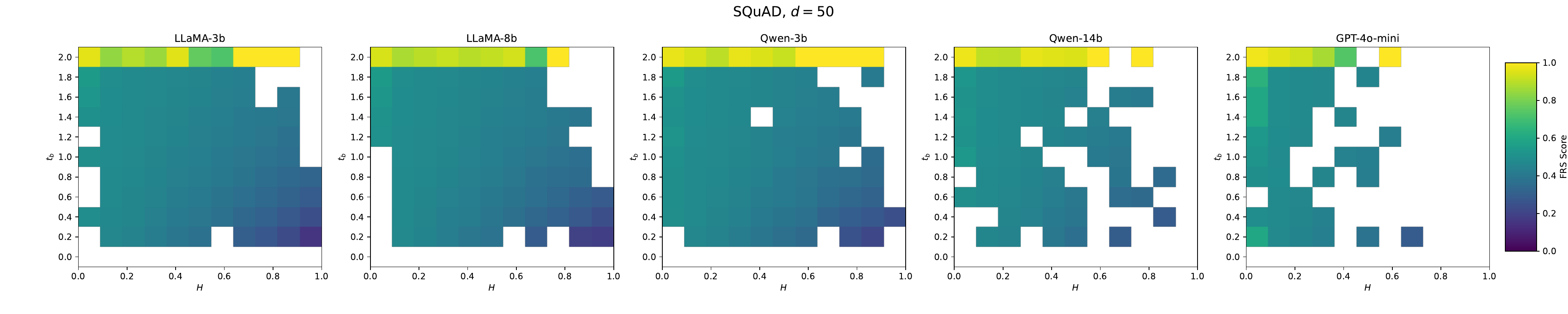}
   \end{minipage}
\caption{SQuAD. FRS scores across models illustrating effects of $d \in \{1,2,5,10,50\}.$}
\label{fig:frs_var_d_squad}
\end{figure*}

\begin{figure*}[htp]

\subfloat{%
  \includegraphics[clip,width=\linewidth]{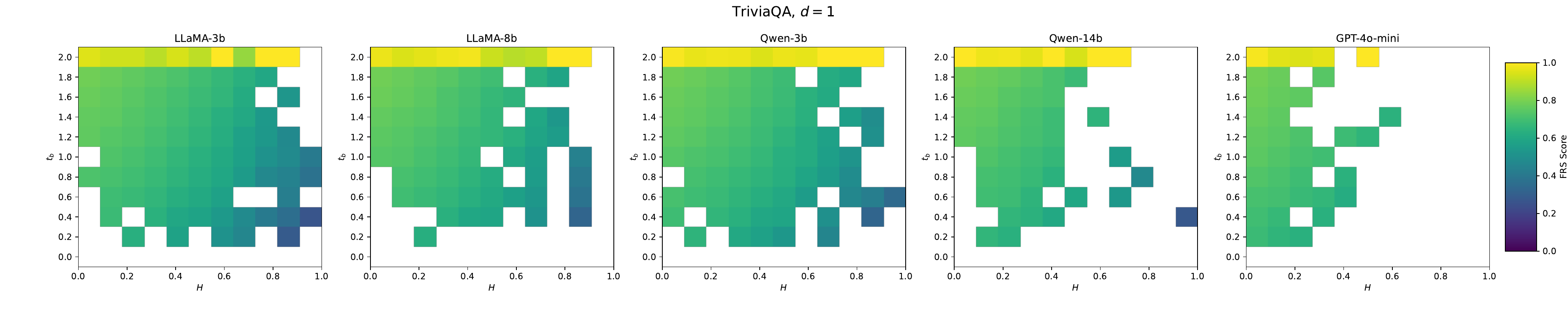}%
}

\subfloat{%
  \includegraphics[clip,width=\linewidth]{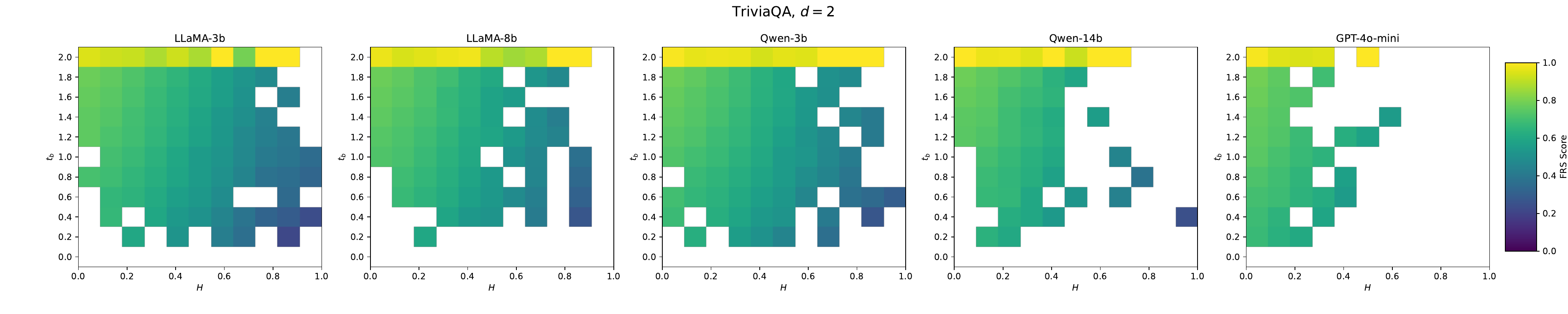}%
}

\subfloat{%
  \includegraphics[clip,width=\linewidth]{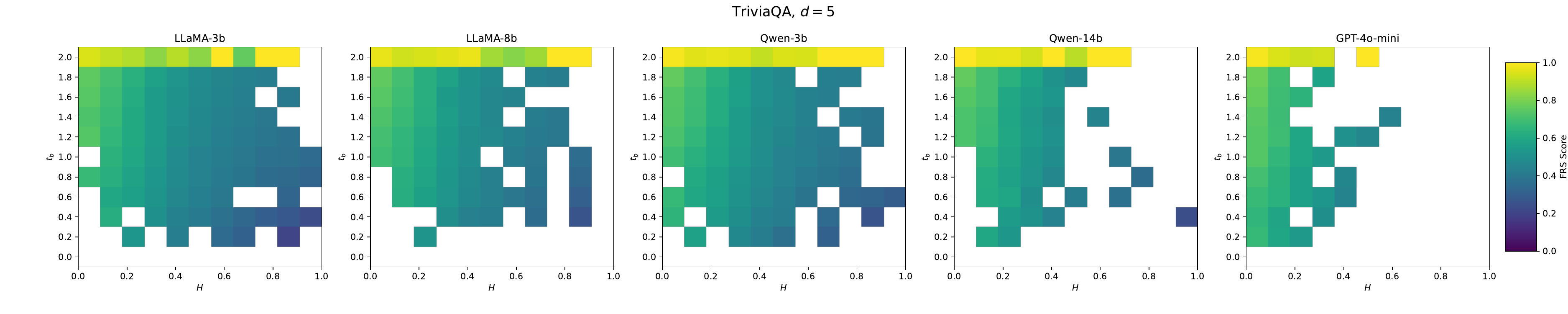}%
}

\subfloat{%
  \includegraphics[clip,width=\linewidth]{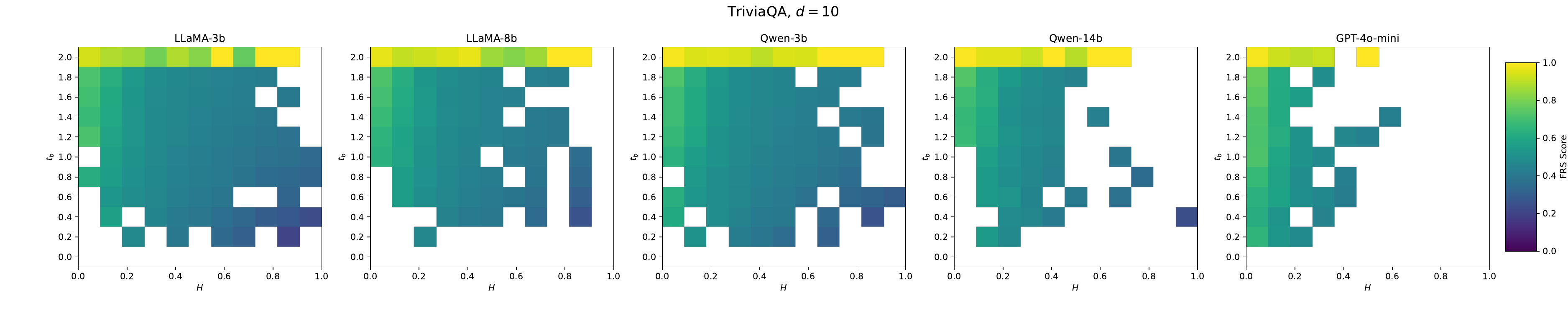}%
}

\subfloat{%
  \includegraphics[clip,width=\linewidth]{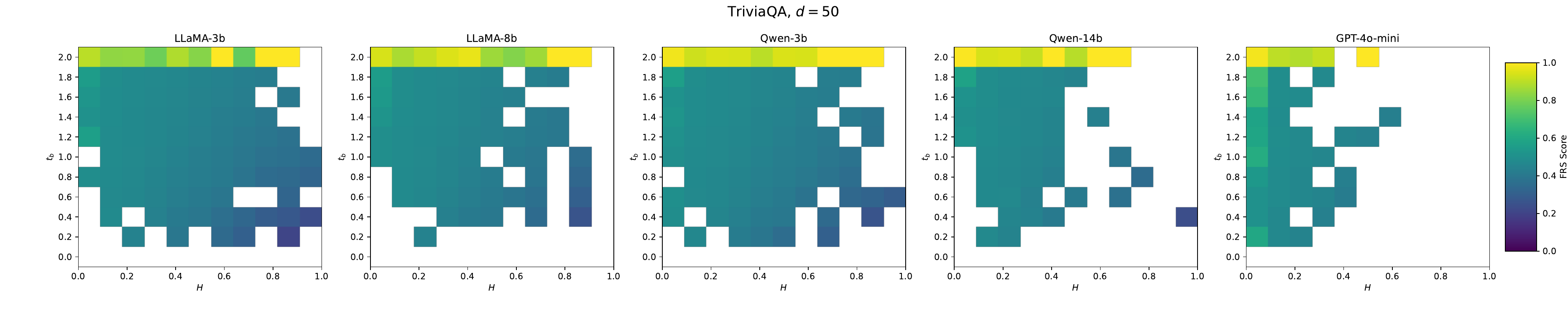}%
}
\caption{TriviaQA. FRS scores across models illustrating effects of $d \in \{1,2,5,10,50\}.$}
\label{fig:frs_var_d_trivia}
\end{figure*}

\begin{figure*}[htp]

\subfloat{%
  \includegraphics[clip,width=\linewidth]{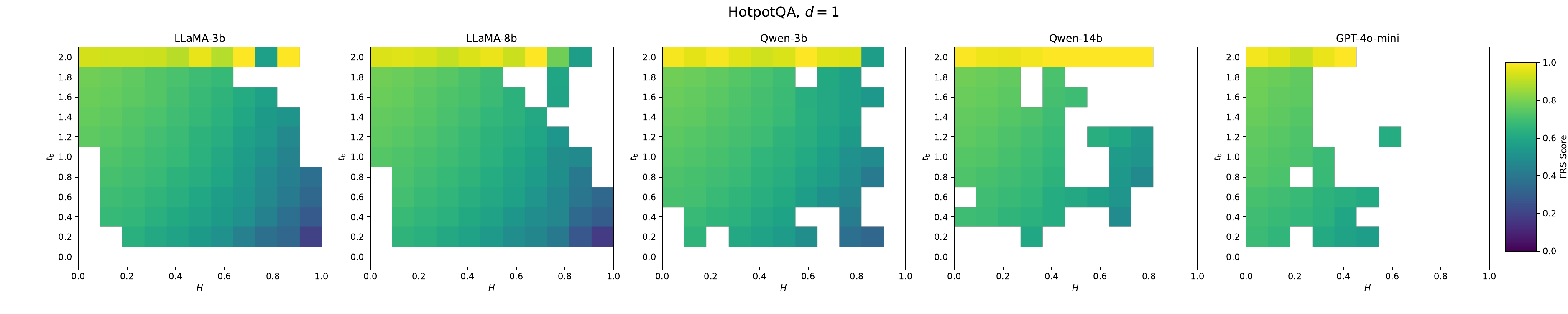}%
}

\subfloat{%
  \includegraphics[clip,width=\linewidth]{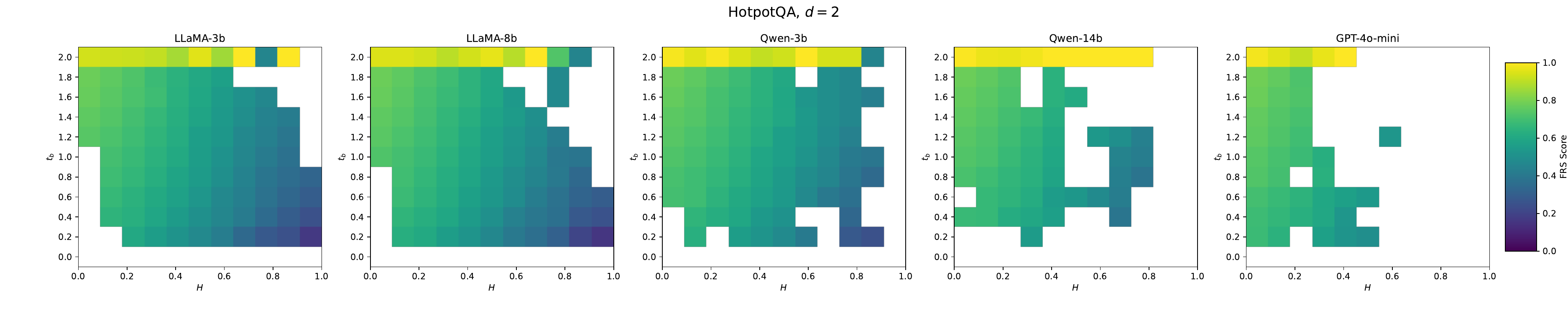}%
}

\subfloat{%
  \includegraphics[clip,width=\linewidth]{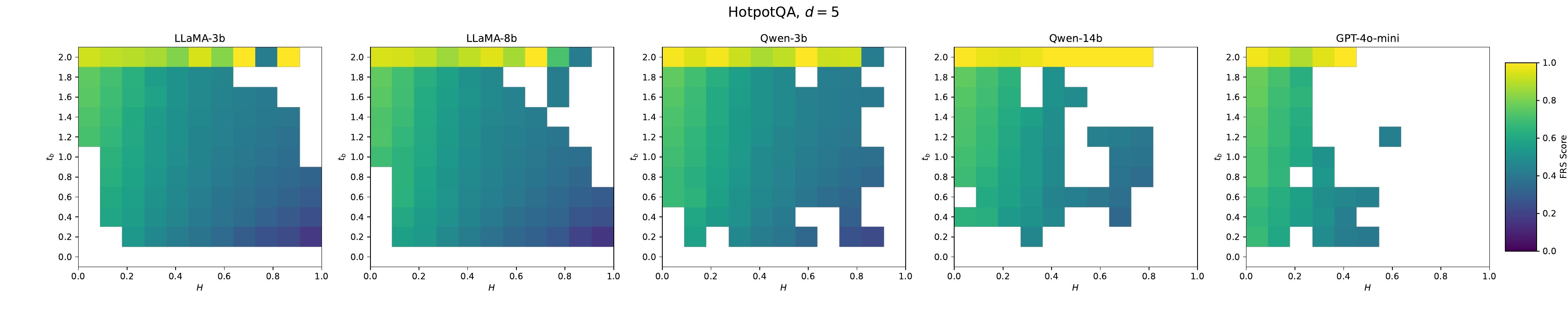}%
}

\subfloat{%
  \includegraphics[clip,width=\linewidth]{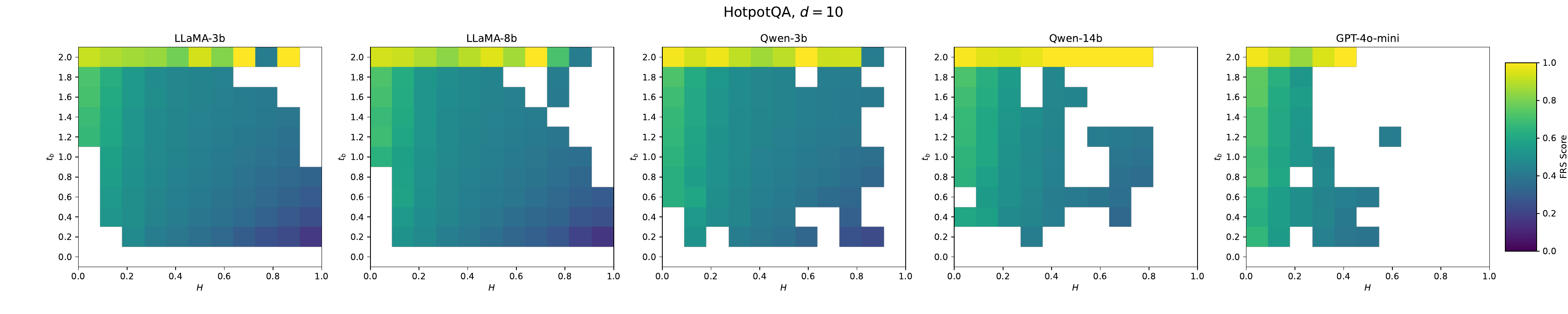}%
}

\subfloat{%
  \includegraphics[clip,width=\linewidth]{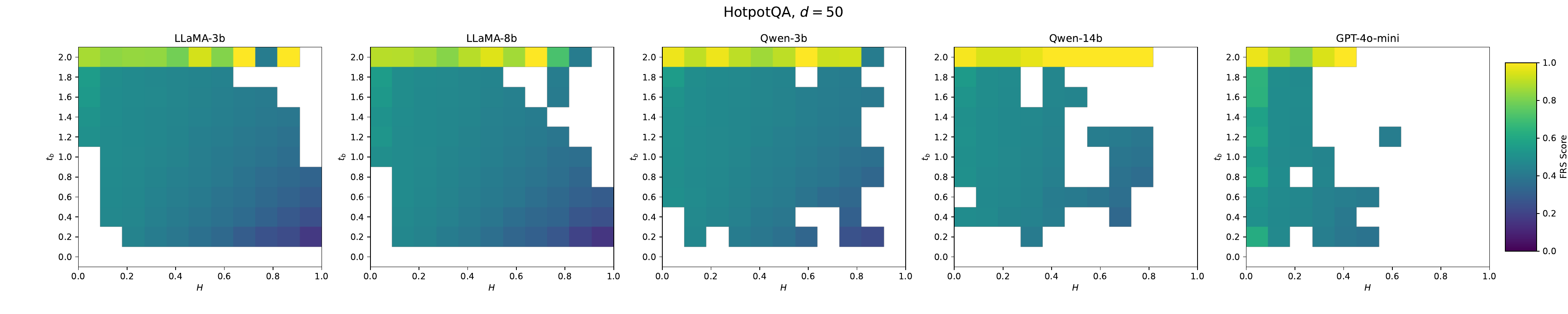}%
}
\caption{HotpotQA. FRS scores across models illustrating effects of $d \in \{1,2,5,10,50\}.$}
\label{fig:frs_var_d_hotpot}
\end{figure*}

In \cref{fig:frs_var_d_squad,fig:frs_var_d_trivia,fig:frs_var_d_hotpot}, we present the FRS scores for all models across various datasets, considering different values of \( d \in \{1,2,5,10,50\} \). As \( d \) increases, the influence of entropy \( H \) on the FRS grows, leading to a steady decline in average FRS scores across all models and datasets. This demonstrates the role of the \( d \) parameter in systematically downscaling the overall robustness score. However, it is important to note that samples without a detected \( t_b \) remain unaffected, consistently maintaining an FRS of 1.

\section{Probability Density Functions of FRS} \label{app:g_probdens}
\begin{figure*}
    \centering
    \begin{minipage}{\linewidth}
        \centering
        \includegraphics[width=\textwidth]{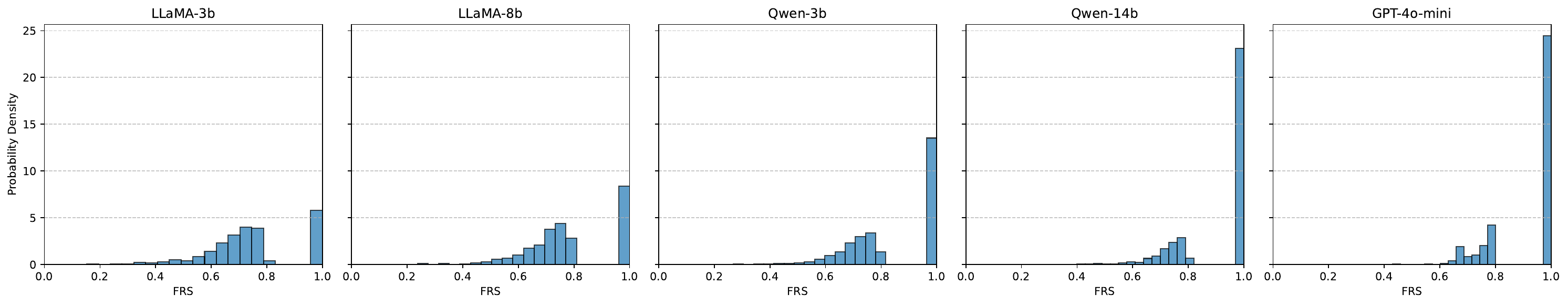}
    \end{minipage}\hfill
    \begin{minipage}{\linewidth}
        \centering
        \includegraphics[width=\textwidth]{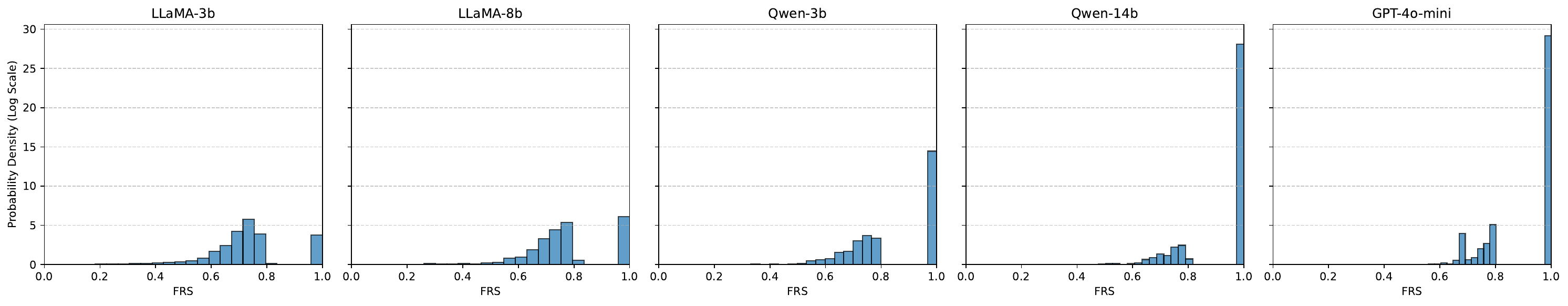}
    \end{minipage}\hfill
    \begin{minipage}{\linewidth}
        \centering
        \includegraphics[width=\textwidth]{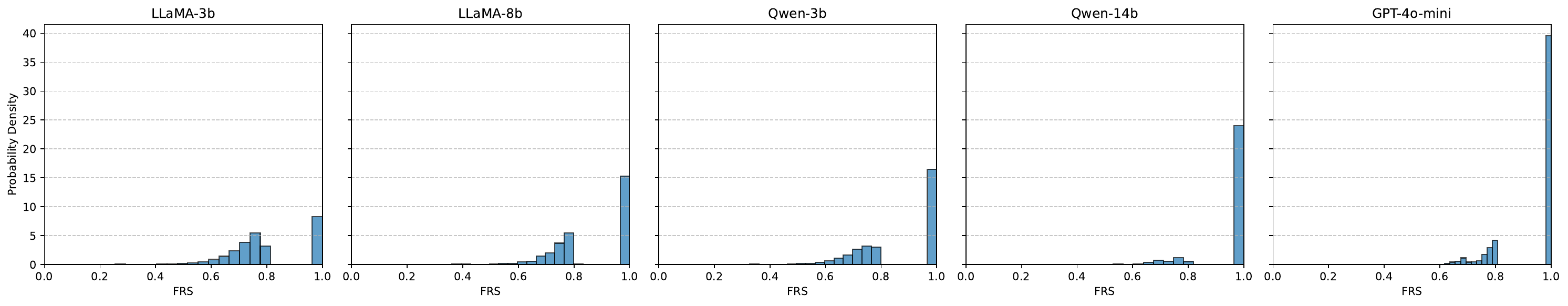}
    \end{minipage}\hfill
    \caption{Probability Density Function of FRS for all models on SQuAD (top), HotpotQA (middle), and TriviaQA (bottom).}
    \label{fig:frs_pdf_hotpot}
\end{figure*}

\cref{fig:frs_pdf_hotpot} presents the Probability Density Functions (PDF) for all models on SQuAD (top), HotpotQA (middle), and TriviaQA (bottom). Notably, Qwen-14b and GPT-4o-mini exhibit comparable robustness levels. As discussed in the main paper, when a model does not break on a question with \( t_b \leq 2 \), we set FRS=1, which explains the observed gap in the x-axis in the interval $[0.8,1)$. A deeper investigation into higher breaking temperatures beyond 2 could help validate the assigned FRS=1 scores or provide more refined estimates, thereby filling the gap in the plots with more accurate robustness scores.

\section{Robust Question Types}
\label{app:h_questions}

\subsection{Training and Dataset}
We fine-tune the base uncased version of DistilBERT, available at \url{https://huggingface.co/distilbert/distilbert-base-uncased}, on the TREC dataset provided at \url{https://huggingface.co/datasets/CogComp/trec}. The training set consists of 5,500 labeled questions, which we use to fine-tune a classifier for categorizing the questions in our own dataset, specifically those associated with the highest and lowest factual robustness scores. Evaluation on the test set (500 questions) yields an accuracy of 96.6\% prior to applying the classifier to our data. 
\cref{tab:training-hyperparams} summarizes the hyperparameters used during training.

\begin{table}[h]
\centering
\caption{Training hyperparameters for DistilBERT fine-tuning on the TREC dataset.}
\resizebox{.6\linewidth}{!}{%
\begin{tabular}{ll}
\toprule
\textbf{Hyperparameter} & \textbf{Value} \\
\midrule
learning rate   & 2e-5 \\
batch size      & 16 \\
epochs & 3 \\
weight decay    & 0.01 \\
\bottomrule
\end{tabular}
}
\label{tab:training-hyperparams}
\end{table}

\noindent The original dataset consists of six entity types: \textit{Numerical, Location, Human, Entity, Abbreviation} and \textit{Description}. In our results, we choose to omit the latter two types, since they are not sufficiently represented in our data, on average making up only 2.4\% and 0.3\%, respectively. Hence, in order to avoid poorly supported claims about the robustness of these entity types, we stick with the other four types that are most characteristic of our data.

\subsection{Examples}
We provide examples of each question type analyzed in~\cref{sec:most_vs_least_robust_facts} for further illustration.

\begin{tcolorbox}[colframe=myblue, colback=myblue!10, sharp corners]
\textbf{Numerical}\\
\textit{Q: In what year did Universal make a film version of Dracula?}\\
\textbf{A:} 1931
\end{tcolorbox}

\noindent\textit{Numerical}-type questions usually ask for a quantity, a date, or similar. The model's response is usually a single number.

\begin{tcolorbox}[colframe=myorange, colback=myorange!10, sharp corners]
\textbf{Location}\\
\textit{Q: The Scorpions came from what country?}\\
\textbf{A:} Germany
\end{tcolorbox}

\noindent\textit{Location}-type questions usually ask for a location, like a country, city, or similar. The model's response is a place.



\begin{tcolorbox}[colframe=myred, colback=myred!10, sharp corners]
\textbf{Human}\\
\textit{Q: Who introduced the first quantized model of the atom?}\\
\textbf{A:} Niels Bohr
\end{tcolorbox}
\noindent\textit{Human}-type questions usually ask for the name of a person or a group of people. The model's response is a name.

\begin{tcolorbox}[colframe=mypurple, colback=mypurple!10, sharp corners]
\textbf{Entity}\\
\textit{Q: What is very similar to Valencian and is considered a variety of the same language?}\\
\textbf{A:} Catalan\\\\
\textit{Q: Guatama discovered the middle path before his what?}\\
\textbf{A:} Enlightenment\\\\
\textit{Q: What color jersey does Bayern Munich wear?}\\
\textbf{A:} Red
\end{tcolorbox}

\noindent\textit{Entity}-type questions are not broadly categorizable. They ask about various facts about the world, which do not typically fall into any of the other categories. For example, they can ask about events, food, language, products. For a full list, see the fine label descriptions on TREC's huggingface dataset card: \url{https://huggingface.co/datasets/CogComp/trec}.


\end{document}